\documentclass[11pt]{article}
\usepackage{amsmath,amsfonts,amssymb,graphicx,amsthm,color}
\usepackage{authblk}
\usepackage[normalem]{ulem}
\usepackage[]{geometry}
\usepackage{colortbl}
\usepackage{booktabs}
\usepackage{xcolor}
\usepackage{amsmath}
\usepackage{algorithm}
\usepackage{algorithmic}
\usepackage{graphicx}
\usepackage{caption}
\usepackage{amssymb}
\usepackage[colorlinks=true, linkcolor=black, citecolor=black, urlcolor=black]{hyperref}
\usepackage{utfsym}
\usepackage{subcaption}

\makeatletter
\DeclareSymbolFont{extraup}{U}{zavm}{m}{n}
\DeclareMathSymbol{\newcheckmark}{\mathalpha}{extraup}{128}%uni2713
\DeclareMathSymbol{\newcrossmark}{\mathalpha}{extraup}{129}%uni2717
\makeatother

\usepackage[title]{appendix}

\theoremstyle{definition}

\definecolor{verdechiaro}{rgb}{0.6,.9,0.6}
\definecolor{mygray}{gray}{0.8}
\usepackage{natbib}

\title{A general framework for adaptive nonparametric dimensionality reduction}

\author[1,2]{Antonio Di Noia}
\author[3]{Federico Ravenda}
\author[2,4]{Antonietta Mira}
\affil[1]{Seminar for Statistics, Department of Mathematics, ETH Zurich}
\affil[2]{Faculty of Economics, Euler Institute, Università della Svizzera italiana}
\affil[3]{Faculty of Informatics, Università della Svizzera italiana}
\affil[4]{Department of Science and High Technology, University of Insubria}

\begin{document}

\maketitle
\date{}

\begin{abstract}
Dimensionality reduction is a fundamental task in modern data science. Several projection methods specifically tailored to take into account the non-linearity of the data via local embeddings have been proposed. Such methods are often based on local neighbourhood structures and require tuning the number of neighbours that define this local structure, and the dimensionality of the lower-dimensional space onto which the data are projected. Such choices critically influence the quality of the resulting embedding. In this paper, we exploit a recently proposed intrinsic dimension estimator which also returns the optimal locally adaptive neighbourhood sizes according to some desirable criteria.
In principle, this adaptive framework can be employed to perform an optimal hyper-parameter tuning of any dimensionality reduction algorithm that relies on local neighbourhood structures.
Numerical experiments on both real-world and simulated datasets show that the proposed method can be used to significantly improve well-known projection methods when employed for various learning tasks, with improvements measurable through both quantitative metrics and the quality of low-dimensional visualizations.%\footnote{Code available at the following Github repository: \href{https://github.com/Fede-stack/Adaptive-nonparametric-dimensionality-reduction}{https://github.com/Fede-stack/Adaptive-nonparametric-dimensionality-reduction}}.
\end{abstract}

\section{Introduction}
\subsection{Background}
The explosion of data production in recent years has created huge amounts of information in many fields, such as medicine, finance, social media, and e-commerce. Such data often has many features or variables, making it high-dimensional. The analysis and visualisation of high-dimensional data is a challenging task because of physical and computational constraints. Dimensionality reduction techniques are essential tools that help make high-dimensional data analysis feasible. These methods aim to reduce the number of features while keeping the relevant information. The main goals of dimensionality reduction are to speed up computations, improve model accuracy, and help data visualisation. A complete and recent survey is given in \cite{ayesha2020overview}.

Low-dimensional representations can be based on both parametric and nonparametric methods; see \cite{gisbrecht2015data}. A famous example of a parametric technique is Principal Component Analysis (PCA) \citep{hotelling1933analysis}, which maps the data into a lower-dimensional space using projections onto linear subspaces. Parametric methods are simple and effective for many data analysis tasks, but may not work well for complex data with strongly nonlinear structures.

Nonparametric approaches are specifically tailored to take into account the strong nonlinearities of the data. A large class of nonparametric methods in statistics is based on Nearest Neighbours (NN). NN methods allow the development of flexible models tailored for several different scopes, that range from density estimation, regression, classification and dimensionality reduction; see \cite{biau2015lectures}.
In nonparametric dimensionality reduction NN methods are adopted to build locally adaptive neighbourhood structures, and the choice of the locally adaptive NN order is fundamental to performing the projection onto a lower-dimensional space \citep{alvarez2011global}.
We mention some celebrated and popular methods: the isometric mapping (Isomap) \citep{tenenbaum2000global}, the Locally Linear Embedding (LLE) \citep{roweis2000nonlinear}, spectral embedding \citep{von2007tutorial}, and Uniform Manifold Approximation and Projection (UMAP) \citep{mcinnes2018umap}
that have been successfully employed to capture nonlinear structures and more complex patterns in the data. However, these methods can be very computationally demanding and hard to tune.

\subsection{Contributions}
In this paper, we introduce a novel nonparametric dimensionality reduction framework that leverages the intrinsic dimension (ID) (see \cite{del2021effective,binnie2025survey} for recent reviews) estimator recently proposed in \cite{di2024beyond}. This estimator jointly estimates the ID of the data and the optimal number of NN, ensuring that the underlying probability distribution of the data is approximately uniform in such a neighbourhood. By automatically identifying the optimal locally adaptive neighbourhood structure for each data point, our method adapts to the local structure of the data, enhancing the performance of NN-based projection methods.
This adaptive approach can be integrated into several well-known nonlinear projection methods, e.g.\ LLE, spectral embedding and UMAP, demonstrating significant improvements in their performance. Our method's adaptability is particularly beneficial in preserving the geometric structure of data, leading to more accurate and meaningful low-dimensional representations.

Through extensive numerical experiments, we evaluate the proposed method across various learning tasks, including classification, clustering, and data visualisation. The results show that our approach not only enhances the accuracy and robustness of nonparametric dimensionality reduction methods but also reduces computational complexity by optimising the neighbourhood selection process.

\subsubsection{Organisation of the paper}
The paper is organised as follows:
In Section~\ref{sec:methods}, we briefly describe the algorithm for estimating the ID and for identifying optimal neighbourhoods for each data point. Successively, we explain how this integrates with nonlinear dimensionality reduction methods such as LLE and Isomap. In Section~\ref{sec:exp}, we show that our framework significantly improves classical nonparametric dimensionality reduction methods for various learning tasks on both real and simulated data. In Section~\ref{sec:disc}, some concluding remarks are offered.

\section{Methods}
\label{sec:methods}

In machine learning, most approaches developed in both supervised and unsupervised settings involve several hyper-parameters to be tuned. These can be set a priori by a domain expert based on the available data type, or can be tuned using Hyper-Parameter Optimisation approaches \citep{yang2020hyperparameter}. Since optimisation algorithms are generally computationally expensive and time-consuming, different techniques have been explored to efficiently search the hyper-parameter space, including Grid Search~\citep{lavalle2004relationship}, Random Search~\citep{bergstra2012random}, and Bayesian Optimization~\citep{snoek2012practicalbayesianoptimizationmachine}.
In particular, in dimensionality reduction algorithms, the nature of hyper-parameters is critical, as it significantly influences how observations are mapped into the reduced space and affects the trade-off between preserving the original local and global structure.
However, it must be pointed out that hyper-parameter tuning in unsupervised approaches poses a significant challenge, as the absence of ground truth labels prevents direct performance validation \citep{yang2020hyperparameter}.
For instance, in LLE and other dimensionality reduction approaches that leverage a neighbourhood structure for the data, the number of components into which the original observations are mapped (\textit{n\_components}) and the number of neighbours to consider (\textit{n\_neighbours}) are the two main hyper-parameters of the algorithms.

In the following, data-driven versions of classical NN-based dimensionality reduction methods are introduced, where hyper-parameters are autonomously estimated by the ID estimator introduced in \cite{di2024beyond} and incorporated within such dimensionality reduction approaches, avoiding further time- and cost-consuming hyper-parameter tuning optimisation. As mentioned, we remark that we adopt such an ID estimation framework, because it naturally adapts to NN-based dimensionality reduction methods.

\subsection{Intrinsic dimension and uniform neighbourhoods}
Let $X_1,\dots,X_n$ be a random sample taking values in $\mathbb{R}^D$. In the following, we use the compact notation $X_{1:n}$ for the sample and $x_{1:n}$ for its realisation. Moreover, suppose that $X_{1:n}$ are sampled in a small neighbourhood of a $d$-dimensional $C^1$ manifold embedded in $\mathbb{R}^D$ with $d<D$; i.e.\ the manifold hypothesis is satisfied.
Let us consider two open balls $B(x,r_A)$ and $B(x,r_B)$ with radii $r_A$ and $r_B$ such that $r_A<r_B$ and centred at the same point on the manifold tangent space. Next, we consider a (spatial) Poisson process on $\mathbb{R}^D$ with intensity $h:\mathbb{R}^D\to \mathbb{R}_+$ and locally homogeneous on $B(x,r_B)$. It follows that the number of realizations belonging to $B(x,r_B)$, denoted by $k_B$, is Poisson distributed with parameter $\rho\mu(B(x,r_B))$, where $\mu$ is the Lebesgue measure on $\mathbb{R}^D$ and $\rho$ is the value attained by the intensity function $h$ on the set $B(x,r_B)$. As a consequence, $k_A|k_B \sim \mathrm{Binomial}(k_B,p)$, where, for a sufficiently small outer ball $B$, $p\approx (r_A/r_B)^d = \tau ^d$, with $d$ corresponding to the ID. Now, let $k_{A,i}$ and $k_{B,i}$ be, respectively, the number of points in the open balls $B(x_i,r_A)$ and $B(x_i,r_B)$, where we do not count $x_i$ since we condition on it. If $k_{A,1:n}$ are independent conditionally on $k_{B,1:n},\tau$, the likelihood of $d$ reads
\begin{equation*}
    L(d)= \prod_{i=1}^n \binom{k_{B,i}}{k_{A,i}}  (\tau^d)^{k_{A,i}} (1-\tau^d)^{k_{B,i}-k_{A,i}}
\end{equation*}
and is maximised at 
\begin{equation}
\label{eq:bide}
    \widehat d = \frac{\log ( (\sum_{i=1}^n k_{A,i}) /(\sum_{i=1}^n k_{B,i}) )}{\log \tau}.
\end{equation}
The estimator $\widehat d$ is called \emph{Binomial ID Estimator} (BIDE). It is immediately noticeable that it depends on the choice of $r_A$ and $r_B$, thus, these radii must be selected. \cite{di2024beyond} proposes an adaptive procedure that exploits a likelihood ratio test to enlarge the neighbourhood of each data point $r_B$ while ensuring that the local homogeneity of the underlying Poisson process still holds. This makes the procedure more efficient and, most importantly, more robust to noise.
The task is turned into the selection of the largest NN of each data point $x_i$ to consider. More precisely, let $r_{i,j}$ be the distance between point $x_i$ and its $j$-th NN and introduce the hyper-spherical shells 
$$v_{i,j}: = \mu(B(x_i,r_{i,j})\setminus B(x_i,r_{i,j-1})) =\Omega_d (r^d_{i,j}-r^d_{i,j-1}),$$ where $\Omega_d = (2\Gamma(3/2))^d/\Gamma(d/2+1)$ is the volume of a unit $d$-dimensional hyper-sphere with $\Gamma$ denoting the Euler's Gamma function. When the underlying Poisson process is homogeneous on $B(x_i, r_{i,j})$ with intensity constantly equal to $\rho_i$, we model $v_{i,1},\dots,v_{i,k}$ as independent and identically distributed draws from an exponential law with rate $\rho_i$. Therefore, the log-likelihood reads
\begin{align}
\label{eq:lik-rho}
    L_{i,k}(\rho_i) := k \log \rho_i -\rho_i V_{i,k},
\end{align}
where $V_{i,k}=\sum^k_{j=1} v_{i,j} $.
The likelihood ratio test compares two models: the first where it is assumed that the intensity of the process at point $x_i$ and its $k+1$ NN are different, say $\rho_i$ and $\rho_i'$, and the second where they are assumed to be equal. Thus, the test statistic 
$$D_{i,k}= -2 \Big(\underset{\rho_i>0}{\max} (L_{i,k}(\rho_i)+L_{k+1,k}(\rho_i))-\underset{\rho_i,\rho'_i>0}{\max} (L_{i,k}(\rho_i)+L_{k+1,k}(\rho'_i))\Big),$$
can be employed to test $H_0: \rho_i=\rho_i'$ against $H_1: \rho_i\neq\rho_i'$. More specifically, the optimal $k$ for each point $i$ is selected as
\begin{equation}
\label{eq:kstar}
    k^*_i = \min \{k : D_{i,k}\geq q_{1-\alpha,1}=:D_\text{thr} \}
\end{equation}
where $q_{1-\alpha,1}$ is the $(1-\alpha)$-quantile of the $\chi_1^2$ distribution.
Note that small values of $\alpha$ imply the identification of larger neighbourhoods. This choice is strongly related to the disentanglement of signal and noise, and it is usually a choice of the researcher.

From \eqref{eq:lik-rho}, it is clear that such a selection procedure depends on $d$, however, \cite{di2024beyond} combined it into an iterative algorithm that starts from an initial estimate of $d$, and improves it through \eqref{eq:kstar} and \eqref{eq:bide} until convergence. Under mild conditions, the algorithm is shown to terminate with small numerical tolerance for $n$ large, and the resulting ID estimator is named Adaptive BIDE (ABIDE). In \cite{di2024beyond} it is proven to be consistent and an expression for its approximate asymptotic variance is provided. The ABIDE workflow outputs the estimated ID and the set of optimal $k_{1:n}^*$ obtained at the convergence of the iterative procedure, which can be regarded as the identified optimal locally adaptive neighbourhood structure. For the purposes of this paper, unless differently specified, we implicitly consider the ID as the nearest integer to the ABIDE estimate and we denote it by $ {d}^*$.

\subsection{Adaptive nonparametric dimensionality reduction}
As already mentioned, the observed data $X_{1:n}=x_{1:n}$ satisfy the manifold hypothesis. As a consequence, there exists a map $f: \mathbb{R}^d\to\mathbb{R}^D$ such that, locally, $f^{-1}(x_{1:n}) = y_{1:n}$, where $y_{1:n}$ is the set of low-dimensional data that we aim to recover. Dimensionality reduction can be seen as the inverse problem of learning the inverse map $g:=f^{-1}$ based on the realisation of a random sample $X_{1:n}$; see \cite{silva2002global}.
The problem is ill-posed, and it can be approached by imposing various restrictions on $f$. A huge literature on different approaches naturally stems from this fact. A large number of dimensionality reduction methods are based on performing the inversion, preserving local properties of the data and of $g$. In the realm of local-preserving methods, very successful procedures that have been proposed are based on local neighbourhood structures induced by NN methods. Notable examples are LLE, Spectral Embedding, and Isomap.
Such a class of methods can be cast into algorithms of the type 
\begin{align}
\label{eq:algo-general}
    y_{1:n}=\widehat g (x_{1:n}; k_{1:n}, d^\text{proj})
\end{align}
where $k_{1:n}$ is the set of local neighbourhoods' sizes expressed in terms of NN orders and $d^{\text{proj}}$ is the dimension of the space where the researcher aims to project the data.
For visualisation purposes, $d^{\text{proj}}=2,3$, while for modelling purposes it can be larger, and the optimal value is the unknown ID $d$. Note that in nonparametric dimensionality reduction methods, $\widehat g$ does not have an explicit expression, with the advantage of being more flexible. In Section~\ref{sec:methods}, we stressed the fact that the ID is fundamentally related to the size of uniform neighbourhoods. Such framework allows us to select $k_{1:n}$ and $d$ in \eqref{eq:algo-general} in an optimal and self-consistent way, i.e.\ for a given dimensionality reduction of the type \eqref{eq:algo-general} we consider its locally adaptive version 
\begin{align}
\label{eq:algo-adaptive}
    y_{1:n}=\widehat {g}^* (x_{1:n}; k^*_{1:n},  d^*)
\end{align}
where $d^*$ is the nearest integer to the ABIDE estimate and $k^*_{1:n}$ is the associated set of sizes for the uniform neighbourhoods of each data point. 

\subsection{Application to LLE: Adaptive LLE}
\begin{figure}[!ht]
    \centering
    \includegraphics[width=1\linewidth]{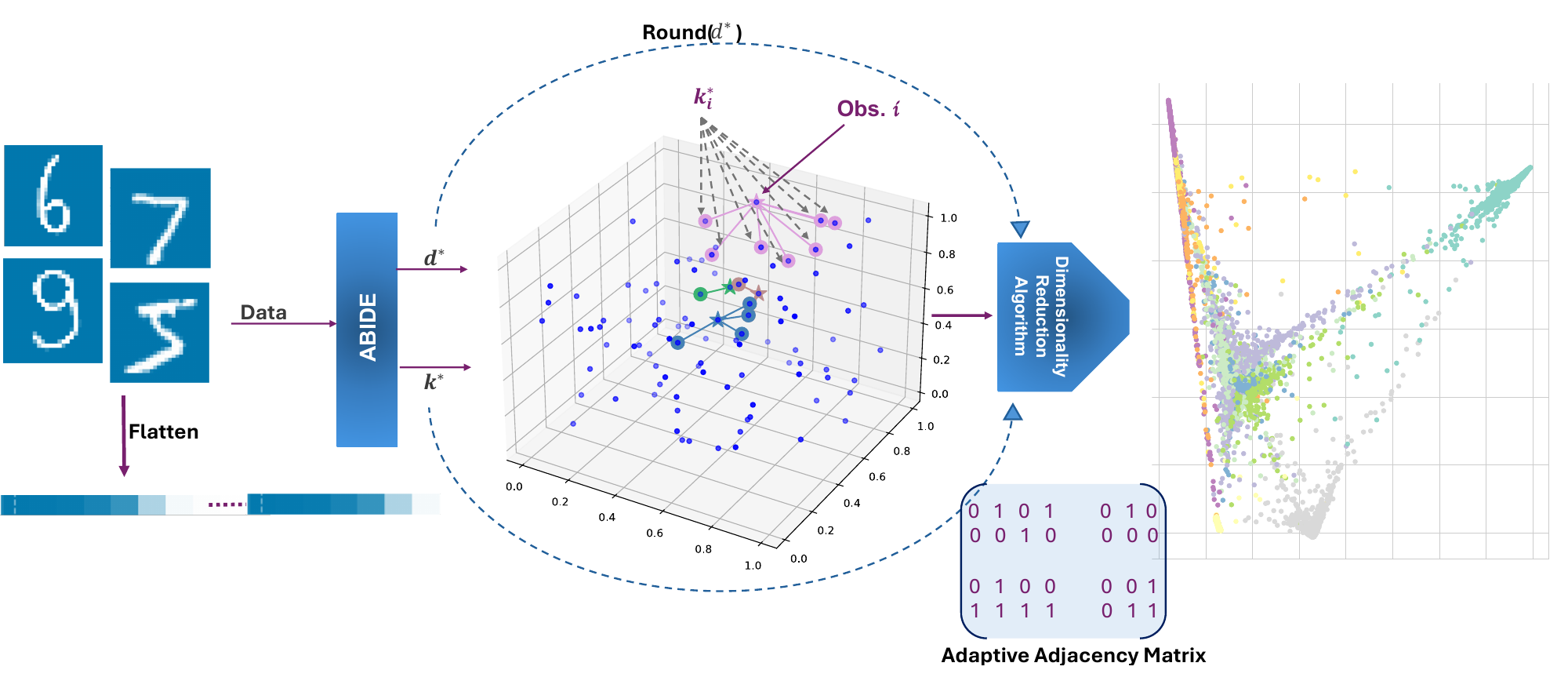}
    \caption{Representation of the main LLE$^*$ steps.}
    \label{llestar}
\end{figure}

Here, we briefly exemplify how the introduced adaptive framework applies to the LLE method (visually represented in Figure \ref{llestar}). Although several variants of LLE have been proposed (e.g.\ \cite{xue2023local,zhang2006mlle}), we adopt the original formulation of \cite{roweis2000nonlinear}; see \cite{ghojogh2020locally} for a review of its variants. 
The (locally) adaptive LLE method indirectly induces the map $\widehat g^*$ in \eqref{eq:algo-adaptive} through the following algorithm composed of three steps: 

In Step 1, given $k_{1:n}^*$, we build a neighbourhood graph where, for each statistical unit $i$, we find its $j$-th nearest neighbour for $j=1,\dots,k_i^*$, according to the Euclidean metric.  
We represent such graph by an adaptive adjacency matrix $A \in \{0,1\}^{n\times n}$, defined as
\begin{align}
\label{eq:adj}
A_{ij} =
\begin{cases}
1 & \text{if } x_j \text{ is among the } k_i^* \text{ nearest neighbours of } x_i,\\
0 & \text{otherwise.}
\end{cases}
\end{align}

In Step 2, we reconstruct each unit $x_i$ via a weighted linear combination of its neighbours. The local reconstruction weights of each point in the $D$-dimensional space, denoted as $\widetilde w_i$, are obtained by solving
\begin{equation}
\label{eq:step1}
\widetilde w_i = \arg \min_{w \in \mathbb{R}^{k_i^*}} \Big\| x_i - \sum_{j: A_{ij}=1} w_j x_j \Big\|_2^2,
\quad \text{subject to } \sum_{j: A_{ij}=1} w_j = 1.
\end{equation}

In Step 3, we define the global weight matrix $W \in \mathbb{R}^{n\times n}$ as
\begin{align}
\label{eq:global-weight}
W_{ij} =
\begin{cases}
(\widetilde w_i)_j & \text{if } A_{ij} = 1,\\
0 & \text{otherwise,}
\end{cases}
\quad i,j=1,\dots,n.
\end{align}

Finally, the $d^*$-dimensional embedding is obtained by solving
\begin{align*}
y_{1:n} = \text{LLE}^*(x_{1:n}; k_{1:n}^*, d^*) 
= &\arg \min_{y_1,\dots,y_n \in \mathbb{R}^{d^*}} \sum_{i=1}^n \Big\| y_i - \sum_{j=1}^n W_{ij} y_j \Big\|_2^2,\\
& \text{subject to} \quad \frac{1}{n} \sum_{i=1}^n y_i y_i^\top = I, \quad \sum_{i=1}^n y_i = 0,
\end{align*}
where $y_i \in \mathbb{R}^{d^*}$ are the embedded points and $I$ is the $d^*\times d^*$ identity matrix.

It is at once apparent that our adaptive version has a strong advantage over the classical LLE because the locally linear reconstructions in Step 1 are much more accurate on local neighbourhoods where the sampling distribution is approximately uniform and, additionally, the size of such neighbourhoods adapts to local variations of the distribution. In Section \ref{sec:exp}, we show that LLE$^*$ outperforms the standard LLE in both supervised and unsupervised learning tasks, as well as in graphical representations.

\section{Numerical experiments}
\label{sec:exp}

\subsection{Datasets}
To evaluate the effectiveness of the proposed method, various datasets with different characteristics are used, allowing for an in-depth exploration across a wide spectrum of data types. In Table~\ref{tabstat}, we refer to some characteristics of each dataset.\\

\noindent\textbf{Iris.} Iris \citep{iris_53} is a simple and widely used dataset in literature, consisting of $n=150$ samples from three species of flowers - Virginia, Versicolor, and Setosa - where each sample contains $D=4$ features. \\
\textbf{MNIST.} MNIST \citep{lecun2010mnist} is a dataset of $28\times 28$ pixel grayscale images of handwritten digits. It contains 10-digit classes (from 0 to 9) with a total of 70000 images. For this analysis, each image is treated as a $D=784$-dimensional vector ($28 \times 28 = 784$ pixels). To better assess the method's performance on a more manageable subset, only the test portion of the dataset is used, which consists of $n=10000$ images. The specific dataset subset used in our experiments is available in the GitHub repository.\\
\textbf{Simulated Manifolds.} We consider $n=5100$ points confined on three different manifolds in the 3D space: a torus, a spiral, and a sphere. Subsequently, we generate $D=20$ Gaussian variables. The first three of these variables are added to the manifold points, while the other 17 become additional (noise) coordinates. In total, there are 5100 observations with 20 variables, of which we already know the true de-noised ID ($d = 3$). This synthetic dataset allows us to test the method's ability in a controlled environment where we already know the labels and the ID of the manifolds. \\
\textbf{News Articles.} A text classification dataset contains $n=2225$ text samples across five categories of documents: politics, sports, technology, entertainment, and business. Starting from the textual content, we generate semantic embeddings using a pre-trained SBERT model \citep{reimers2019sentencebertsentenceembeddingsusing}, specifically the \texttt{sentence-t5-base} model. These generated $D=768$-dimensional vectors become the input for the dimensionality reduction approach.

\begin{table}[!ht]
\centering
 \resizebox{\linewidth}{!}{%
\begin{tabular}{lrrrrrrrr}
\hline
\textbf{Dataset} & \textbf{Synthetic} & \textbf{Type} & \textbf{\# of Vars ($D$)} & \textbf{\# of Obs ($n$)} & \textbf{\# of Labels}  & %\textbf{ID Estimated (
$d^*$%)} 
& \textbf{std}($d^*$) & $\Delta_{time}$\\
\hline
\textbf{Iris} & $\newcrossmark$ & Tabular & 4 & 150 & 3 &  2.55 & 0.06 & 0.02\\  
\hline
\textbf{MNIST} & $\newcrossmark$ & Image & 784 & 10000 & 10  & 11.56 & 0.05 &1.01\\
\hline
\textbf{Manifolds} & $\newcheckmark$ & Tabular &20 & 5100  & 3  & 3.30 & 0.01 & 0.23\\
\hline
\textbf{News Articles} & $\newcrossmark$ & Text &768 & 2225 & 5 &  10.94 & 0.09 & 0.13\\
\hline
\end{tabular}}
\normalsize
\caption{Main characteristics of the datasets used, namely the data type, number of variables, number of observations, number of labels, the  estimated ID $d^*$, the approximate standard deviation of the ID estimator, and the ABIDE's execution time (measured in seconds), which represents the additional computational cost introduced by the adaptive mechanism compared to its non-adaptive counterpart.
}
\label{tabstat}
\end{table}
In Table \ref{tabstat}, we also reported $\Delta_{time}$, which represents the execution time of ABIDE (measured in seconds) required to estimate the optimal number of neighbours and the intrinsic dimension.
The value $\Delta_{time}$ therefore represents the additional computational cost, measured on a Mac equipped with an M2 chip, introduced by the adaptive versions relative to the non-adaptive ones.

\subsection{Unsupervised learning}

In the following subsections, we evaluate the quality of the representations generated in a fully unsupervised context. The adaptive extension we propose can be used both in data preprocessing contexts, where the ID estimation and neighbourhood structure are used jointly to map the original data into a low-dimensional space, and in a data visualization scenario, fixing the dimension of the target space, leveraging only the adaptive structure inferred by the ID estimator.

\subsubsection{Qualitative Evaluation}

To qualitatively evaluate the representations of the adaptive model, we map the observations into a 2D space for the MNIST and News Articles datasets, and into a 3D space for the Manifolds dataset, of which we know the true ID.

\begin{figure}[!ht]
    \centering
    \includegraphics[width=1\linewidth]{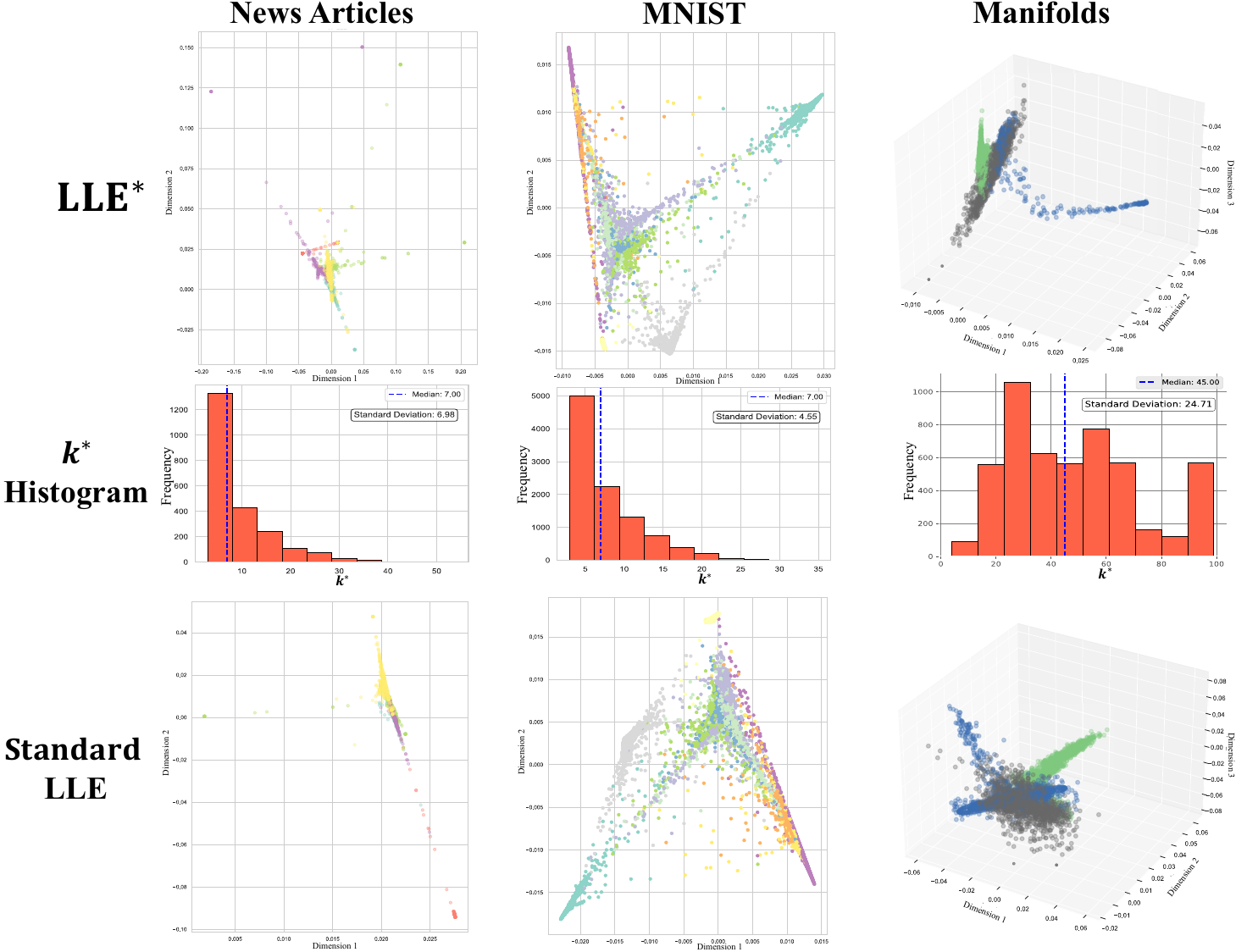}
    \caption{Data visualisation in the reduced space using LLE$^*$ (first row) and the standard version (third row). The second row shows histograms of $k^*$, with summary statistics displayed for median values and standard deviation.}
    \label{multipleviz}
\end{figure}

In Figure \ref{multipleviz}, we observe the representations generated by LLE$^*$, the histograms related to the estimated values of $k^*$, and the representations generated by LLE using the median value of the $k^*$ distribution as the fixed number of neighbours (\textit{n\_neighbours}).
We observe that in the case of MNIST, the two representations are very similar to each other. We anticipate that this similarity will carry over to how the two approaches perform when clustering is applied to the reduced data obtained with the adaptive and non-adaptive LLE methods.
Concerning News Articles, we observe that LLE$^*$ produces more compact clusters with well-identified boundaries, while standard LLE shows significant overlap that obscures the underlying data structure.
In Manifolds, the representation obtained with LLE$^*$ improves the identification of clusters.
In this case, we also examine the  $k^*$
histograms, which differ significantly from the previous ones. While the previous distributions were unimodal and positively skewed with low variability, here the standard deviation is much larger, indicating that the data density is strongly non-uniform. As a result, the LLE$^*$ adaptive approach is better suited to capture local variations.

\subsubsection{Quantitative Evaluation} \label{quant}
To quantitatively assess the quality of representations generated by adaptive methods compared to their non-adaptive counterparts, we compare the results obtained from clustering algorithms applied to the low-dimensional representations. 
We use a simple and widely used clustering approach, K-Means~\citep{lloyd1982least}, to group data into as many clusters as the number of labels in the considered datasets. Once we obtain the output from the clustering approach, we compare the predictions with the ground truths, as reported in the original datasets.
We evaluate the clustering performance using metrics adapted for scenarios where clustering is performed with known true labels. The metrics used are Adjusted Rand Index (ARI) \citep{hubert1985comparing}, Completeness, Homogeneity, and V-measure \citep{rosenberg2007v}. 
In particular, the ARI measures clustering similarity, ranging from -1 to 1, with 1 indicating perfect agreement with true labels. Completeness assesses if members of the same true class are in the same cluster, while Homogeneity evaluates how pure or homogeneous each cluster is with respect to a single class label. The V-measure balances homogeneity and completeness, calculated as their harmonic mean. All these metrics range from 0 to 1 (except ARI), with 1 being the optimal score.

As mentioned, the proposed approach allows for autonomously optimising the hyper-parameters by exploiting the locally adaptive neighbourhood structure of the data, independently finding the best hyper-parameters for the number of neighbours (\textit{n\_neighbours}) and the dimension on which to map the original variables (\textit{n\_components}).

\begin{table}[!ht]
\centering
\begin{tabular}{lcccc}
\hline
\textbf{Models} & \textbf{ARI} & \textbf{Homogeneity} & \textbf{Completeness} & \textbf{V-Measure} \\
\hline
\multicolumn{5}{|c|}{\cellcolor{mygray}\textcolor{black}{\textbf{MNIST}}} \\
\hline
\textbf{LLE}$^*$ & \textbf{0.586} & \textbf{0.712}  & \textbf{0.789} & \textbf{0.749}\\
\hline
\textbf{LLE} & 0.465 & 0.578 & 0.739 & 0.649 \\
\hline
\textbf{LLE modified} & 0.352 & 0.543 & 0.691 & 0.608 \\
\hline
\multicolumn{5}{|c|}{\cellcolor{mygray}\textcolor{black}{\textbf{News Articles}}} \\
\hline
\textbf{LLE}$^*$ & \textbf{0.281} & \textbf{0.411}  & \textbf{0.664} & \textbf{0.507}\\
\hline
\textbf{LLE} & 0.206 & 0.292 & 0.497 & 0.369 \\
\hline
\textbf{LLE modified} & 0.001 & 0.005 & 0.227 & 0.010 \\
\hline
\multicolumn{5}{|c|}{\cellcolor{mygray}\textcolor{black}{\textbf{Manifolds}}} \\
\hline
\textbf{LLE}$^*$ &\textbf{0.719}  & \textbf{0.689} & \textbf{0.695} &  \textbf{0.692}\\
\hline
\textbf{LLE} & 0.318 & 0.422 & 0.582 & 0.489 \\
\hline
\textbf{LLE modified} & 0.101 & 0.201 & 0.394 & 0.266 \\
\hline
\end{tabular}
\normalsize
\caption{Clustering performance metrics based on MNIST, News Articles, and Manifolds dataset. We compared $\text{LLE}^*$ with the default configuration of sklearn of the standard LLE implementation and the modified one. Best results for each metric are highlighted in \textbf{bold}.}
\label{llemod}
\end{table}

In Table \ref{llemod}, we report the results of LLE* together with those obtained using the default hyper-parameter configuration of the standard LLE implementation from the \texttt{scikit-learn} Python package and the modified version of LLE \citep{zhang2006mlle} . This comparison is motivated by the 
scenario in which a practitioner, unaware of the most suitable hyper-parameter settings, would typically rely on the default configuration provided by standard software packages.

\begin{figure}[!ht]
    \centering
    \includegraphics[width=1\linewidth]{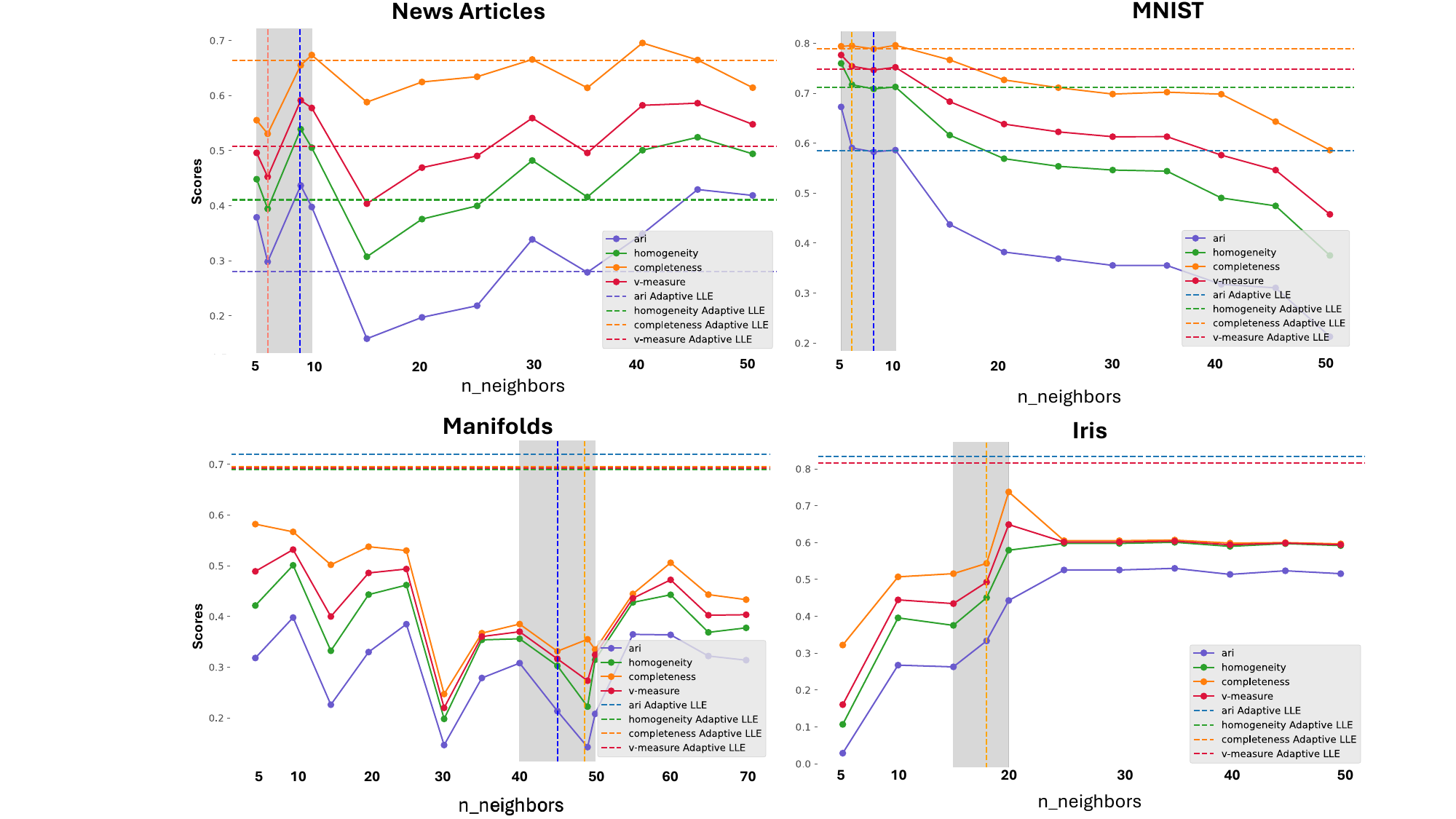}
    \caption{Visualisation of performance metrics of ARI, Homogeneity, Completeness, and V-Measure as the number of considered neighbours varies, fixing the dimension of the target space $d^*$, returned by ABIDE, for the 4 datasets considered. Horizontally, in dashed lines, we report the metrics of LLE$^*$. Vertically, the median (in orange) and the mean (in blue) of $k^*$ are shown. The grey shaded area represents an arbitrary neighbourhood around these summary measures (mean and median) to highlight the proximity of the adaptive results to these reference values. With respect to the Iris dataset, the mean and median coincide.}
    \label{alltog}
\end{figure}

In Figure \ref{alltog}, we compare the performance of LLE$^*$ (horizontal dashed lines) with the performance of LLE as the number of neighbours varies (continuous lines) for all 4 chosen metrics. We create a combination of possible hyper-parameter values explored through grid search. This allows us to evaluate how LLE$^*$ behaves compared to different choices of numbers of neighbours (\textit{n\_components} is fixed and equal to $d^*$). We explore neighbours from 5 to 50 with a step size of 5, except for Manifolds, where, since the average $k^*$ is close to 50, we extend the exploration up to 70 neighbours. Vertical dashed lines are also drawn to highlight the mean (blue) and the median (orange) of the $k^*$ distribution.
We observe that in most cases, the results obtained by varying the number of neighbours are smaller, and thus worse, or close to the results obtained with  $\text{LLE}^*$.

Another interesting observation is that when we fix the number of neighbours as the mean or the median - or we consider a neighbourhood around these summary statistics (represented by the gray area in Figure \ref{alltog})- of the $k^*$ distribution, the results obtained for some metrics are comparable to those obtained by using $k^*$ for the MNIST, Iris, and News Articles datasets. Regarding the Manifolds dataset, this trend is not visible: as previously discussed, looking at Figure \ref{multipleviz}, we can see how the distribution of $k^*$ is much more variable compared to that of the other datasets, which indicates that the data density varies significantly across different regions of the manifold. Therefore, if we take the median of $k^*$, we possibly introduce systematic biases due to sub-regions where the density exhibits strong irregularities.

These insights allow us to make some remarks on how to handle the possible modelling scenarios: 
\begin{itemize}
    \item When the data density of points varies considerably, the locally adaptive structure enables us to effectively describe the local neighbourhoods of observations, avoiding the introduction of bias during the projection phase into the low-dimensional space.
    \item For data with more regular density, on the other hand, imputing a fixed $k$ as the median $k^*$ already leads to acceptable results. However, since the data distribution is unknown, we recommend always using LLE$^*$ because it leads to accurate and robust results independently of the roughness of the data density.
\end{itemize}

\begin{figure}[!ht]
    \centering
    \includegraphics[width=1\linewidth]{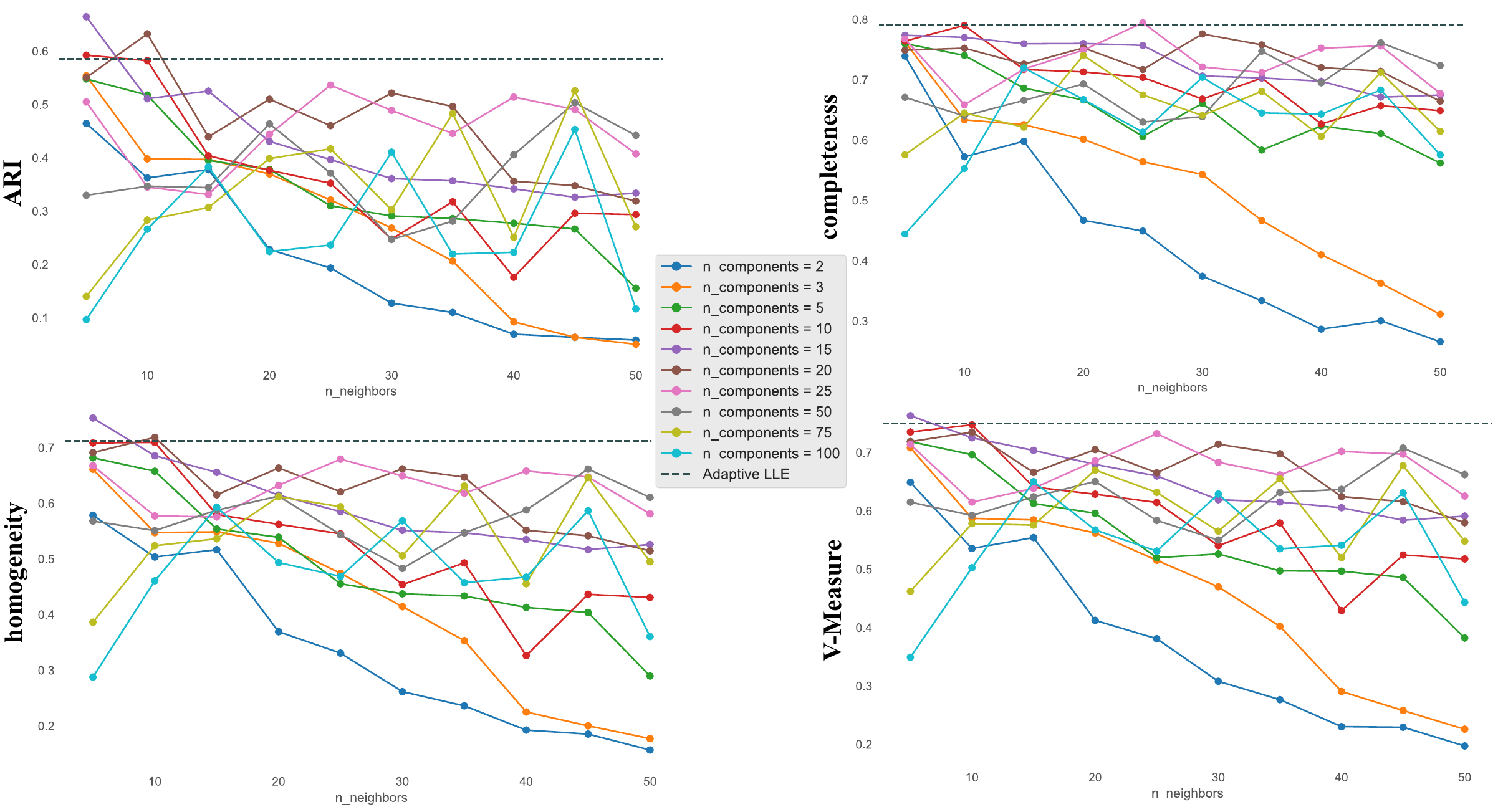}
    \caption{Visualisation of the performance of the 4 considered metrics ARI, Homogeneity, Completeness, and V-Measure for different choices of \textit{n\_components} and \textit{n\_neighbours} in LLE on MNIST dataset. The horizontal dashed lines represent the results of LLE$^*$.}
    \label{mnistopt}
\end{figure}

For the sake of completeness, in Figure \ref{mnistopt}, we observe how performance varies as the two hyper-parameters \textit{n\_components} and \textit{n\_neighbours} change for the MNIST dataset. Again, we observe that the choice of the adaptive method leads to better results for almost all the possible configurations of hyper-parameters explored and for all the considered metrics. 
For some metrics, the best results are obtained with combinations of n\_components = 10 and 15 and n\_neighbours = 5, values that are comparable to those obtained with the ID estimate ($d^*$ = 11.56) and the median number of $k^*$ of 7.

\subsubsection{Beyond LLE: Application to spectral methods}
The adaptive framework introduced for LLE$^*$ naturally extends to many other NN based dimensionality reduction methods.
As a proof-of-concept, we test the performance of our adaptive paradigm in Spectral Clustering (SC) \citep{von2007tutorial}.
In SC, a similarity graph is constructed using a fixed number of NNs, followed by eigendecomposition of the graph Laplacian and $k$-means clustering on the resulting spectral embedding. 
We develop an adaptive version of Spectral Clustering (SC$^*$) that leverages the ABIDE estimator to automatically determine both the optimal neighbourhood structure and the embedding dimensionality for spectral graph construction.
SC$^*$ replaces the fixed neighbourhood parameter with the data-driven $k^*$ values obtained from ABIDE, while using $d^*$ to determine the spectral embedding dimensionality.
The most basic adaptive version of SC can be obtained by simply plugging the adaptive adjacency matrix $A$ introduced in \eqref{eq:adj} of Section \ref{sec:methods} into the standard spectral embedding workflow. For comparison, SC corresponds to the \texttt{scikit-learn} implementation configured to build the similarity graph using nearest neighbours instead of the default RBF kernel, while keeping all the other hyper-parameters at their default values.

\begin{table}[!ht]
\centering
\begin{tabular}{lcccc}
\hline
\textbf{Models} & \textbf{ARI} & \textbf{Homogeneity} & \textbf{Completeness} & \textbf{V-Measure} \\
\hline
\multicolumn{5}{|c|}{\cellcolor{mygray}\textcolor{black}{\textbf{MNIST}}} \\
\hline
\textbf{SC}$^*$ & \textbf{0.589} & \textbf{0.725}  & \textbf{0.845} & \textbf{0.780}\\
\hline
\textbf{SC} & 0.563 & 0.701 & 0.722 & 0.712 \\
\hline
\multicolumn{5}{|c|}{\cellcolor{mygray}\textcolor{black}{\textbf{News Articles}}} \\
\hline
\textbf{SC}$^*$ & \textbf{0.539}& \textbf{0.585}& 0.694  &\textbf{ 0.635} \\
\hline
\textbf{SC} &  0.447& 0.557 & \textbf{0.704} & 0.622 \\
\hline
\multicolumn{5}{|c|}{\cellcolor{mygray}\textcolor{black}{\textbf{Manifolds}}} \\
\hline
\textbf{SC}$^*$ & \textbf{0.673} & \textbf{0.694}  & \textbf{0.713} & \textbf{0.703} \\
\hline
\textbf{SC} & 0.263 & 0.367 & 0.533 & 0.435\\
\hline
\end{tabular}
\normalsize
\caption{Clustering performance metrics based on MNIST, News Articles, and Manifolds dataset. We compared $\text{SC}^*$ with the default version of sklearn. Best results for each metric are highlighted in \textbf{bold}.}
\label{tab:SC}
\end{table}

Experimental evaluation on MNIST, News Articles, and Manifolds datasets in Table \ref{tab:SC} demonstrates the effectiveness of this adaptive approach. SC$^*$ achieves substantial improvements over standard SC on MNIST and News Articles, and particularly dramatic gains on the Manifolds dataset, where the non-uniform density distribution makes the locally adaptive neighbourhood selection crucial. 
%({\href{https://scikit-learn.org/stable/modules/generated/sklearn.cluster.SpectralClustering.html}%{\textcolor{blue}{https://sklearn.cluster.SpectralClustering.html}})

\subsubsection{Beyond LLE: Application to UMAP}
UMAP (Uniform Manifold Approximation and Projection) is a nonparametric dimensionality reduction method based on manifold learning and topological data analysis \citep{mcinnes2018umap}. The algorithm constructs a fuzzy topological representation of the data in the high-dimensional space and then optimizes a low-dimensional embedding to preserve this structure. A critical hyper-parameter in UMAP is the number of nearest neighbours (n\_neighbours) used to construct the local neighbourhood graph, which directly influences the balance between preserving local and global structure.

We introduce a simple adaptive version of UMAP, denoted as UMAP$^*$, that leverages the locally adaptive neighbourhood structure $k^*_{1:n}$ obtained from the ABIDE estimator.
% Similarly to SC$^*$, this is done by inputting the adaptive adjacency matrix $A$ introduced in Section \ref{sec:methods}.
Adopting the notations introduced in Section \ref{sec:methods} and following the standard UMAP workflow \citep{mcinnes2018umap}, we introduce local scaling parameters $\sigma_i$ such that
$$
\sum_{j: A_{ij}=1} \exp\Big(-\frac{\max(0,\|x_i-x_j\|_2-\rho_i)}{\sigma_i}\Big) = \log_2 k_i^*,
$$
where $\rho_i$ is the distance 
to first non-coincident nearest neighbour. Next, the local fuzzy weight vectors are computed as
$$
 (\widetilde w_i)_j = \exp\Big(-\frac{\max(0,\|x_i-x_j\|_2-\rho_i)}{\sigma_i}\Big), \quad \text{ for all } j: A_{ij}=1,
$$
and the global weighted adjacency matrix is obatined as in \eqref{eq:global-weight}:
$$
W_{ij} =
\begin{cases}
(\widetilde w_i)_j & \text{if } A_{ij}=1,\\
0 & \text{otherwise,}
\end{cases}
\quad i,j=1,\dots,n.
$$
The sparse weight matrix $W$ defines the fuzzy simplicial set used by the UMAP workflow. Thus, following the standard UMAP procedure, $W$ is symmetrized,
and its symmetrized version defines the fuzzy topological representation of the data in the high-dimensional space, where the locally varying neighbourhood sizes $k^*_{1:n}$ naturally adapt to regions of varying data density.
Finally, as in UMAP, UMAP$^*$ constructs a $d^*$-dimensional representation by minimizing the fuzzy set cross-entropy between the high-dimensional and low-dimensional fuzzy representations.

\begin{table}[!ht]
\centering
\begin{tabular}{lcccc}
\hline
\textbf{Models} & \textbf{ARI} & \textbf{Homogeneity} & \textbf{Completeness} & \textbf{V-Measure} \\
\hline
\multicolumn{5}{|c|}{\cellcolor{mygray}\textcolor{black}{\textbf{MNIST}}} \\
\hline
\textbf{UMAP}$^*$ & \textbf{0.753} & \textbf{0.818}  & \textbf{0.840} & \textbf{0.829}\\
\hline
\textbf{UMAP} & 0.748 & 0.810 & 0.831 & 0.813 \\
\hline
\multicolumn{5}{|c|}{\cellcolor{mygray}\textcolor{black}{\textbf{News Articles}}} \\
\hline
\textbf{UMAP}$^*$ & \textbf{0.641}& \textbf{0.729}& 0.775  & 0.760\\
\hline
\textbf{UMAP} &  0.618& 0.727 & \textbf{0.796} & 0.760 \\
\hline
\multicolumn{5}{|c|}{\cellcolor{mygray}\textcolor{black}{\textbf{Manifolds}}} \\
\hline
\textbf{UMAP}$^*$ & \textbf{0.788} & \textbf{0.768}  & \textbf{0.777} & \textbf{0.773} \\
\hline
\textbf{UMAP} & 0.447& 0.469 & 0.491 & 0.480\\
\hline
\end{tabular}
\normalsize
\caption{Clustering performance metrics based on MNIST, News Articles, and Manifolds dataset. We compared $\text{UMAP}^*$ with the default version of \texttt{uwot} package. Best results for each metric are highlighted in \textbf{bold}.}
\label{UMAPA}
\end{table}

Based on Table \ref{UMAPA}, we observe that the adaptive model (UMAP$^*$) achieves better performance on the considered datasets compared to the default hyper-parameter configuration of the \texttt{uwot} package. For MNIST, UMAP$^*$ consistently outperforms standard UMAP across all the considered evaluation metrics, demonstrating that the locally adaptive neighbourhood structure better captures the local variations of the underlying data distribution. The improvement is even more pronounced for the Manifolds dataset, where UMAP$^*$ achieves markedly higher clustering accuracy and structure preservation. This result highlights once again the advantage of using locally adaptive neighbourhoods when the data density varies strongly across regions of the manifold, allowing the model to flexibly adjust to local variations and avoid distortions in the low-dimensional embedding.
For the News Articles dataset, the performance gap between the two variants is less pronounced, with UMAP$^*$ showing a slight improvement in ARI and Homogeneity, while maintaining comparable V-Measure scores.

\subsection{Supervised learning}

For supervised learning tasks, we must address the challenge of extending our adaptive framework to out-of-sample points not present during the training phase. We develop a local reconstruction approach that maintains the adaptive neighbourhood structure for test data. For each test point, ${x}_{test}$, we temporarily augment the training data including it, estimate its optimal neighbourhood size, ${k}^*_{test}$ using the intrinsic dimension learned during training, and compute locally adaptive reconstruction weights based on its $k_{test}^*$ nearest neighbours. The test point is then projected into the embedding space using a weighted combination of its neighbours' embeddings. See Algorithm \ref{alg} for the detailed procedure.

\begin{algorithm}[H]
{
\caption{Out-of-Sample LLE$^*$ for Supervised Learning}
\label{alg:out_of_sample}
\begin{algorithmic}[1]
\REQUIRE Training data ${X}_{\text{train}} \in \mathbb{R}^{n \times D}$, test point ${x}_{\text{test}} \in \mathbb{R}^D$, trained embedding ${Y}_{\text{train}} \in \mathbb{R}^{n \times d^*}$, dimension estimate $d^*$
\ENSURE Test point embedding ${y}_{\text{test}} \in \mathbb{R}^{d^*}$

\textbf{Augmented Dataset Construction:}
\STATE Construct augmented dataset $X_{\text{aug}} = \begin{bmatrix}   X_{\text{train}} \\ x_{\text{test}} \end{bmatrix} \in \mathbb{R}^{(n+1)\times D}$, initialize data with ${X}_{\text{aug}}$

\textbf{Dimension and Neighbourhood Estimation:}
\STATE Apply ABIDE: set intrinsic dimension to $d^*$ and the optimal neighbourhood for ${x}_{\text{test}}$ as the $k^*_{\text{test}}$ NN of ${x}_{\text{test}}$ in ${X}_{\text{train}}$ denoted as ${x}_{\text{test}}^{(1)},\ldots, {x}_{\text{test}}^{(k^*_{\text{test}})}$\\
\textbf{Local Linear Reconstruction:}
\STATE Find weights ${w}$ that minimize $\|{x}_{\text{test}} - \sum_{j=1}^{k^*_{\text{test}}} w_j {x}_{\text{test}}^{(j)}\|_2^2$ subject to $\sum_{j=1}^{k^*_{\text{test}}} w_j = 1$

\textbf{Embedding Projection:}
\STATE Project into embedding space: ${y}_{\text{test}} = \sum_{j=1}^{k^*_{\text{test}}} w_j {y}_{\text{test}}^{(j)}$ where ${y}_{\text{test}}^{(1)},\ldots, {y}_{\text{test}}^{(k^*_{\text{test}})}$ are the projections of ${x}_{\text{test}}^{(1)},\ldots, {x}_{\text{test}}^{(k^*_{\text{test}})}$ contained in $Y_{\text{train}}$
\RETURN ${y}_{\text{test}}$
\end{algorithmic}
\label{alg}}
\end{algorithm}

In a supervised context, the projections obtained as output from dimensionality reduction algorithms are often fed to supervised models. This is done to avoid possible complications due to the high dimensionality of the data, e.g., the curse of dimensionality \citep{verleysen2005curse}, to avoid overfitting, to decrease the computational burden of high-dimensional datasets, and ultimately to improve model performance \citep{coelho2022review,huang2024deep}.

To evaluate the quality of the representations in a supervised context, we use the MNIST, Manifolds, and News Articles datasets and assess the performance in terms of accuracy and F1 score. To do this, we use $m$-fold cross-validation, with $m = 3$, where a Logistic Regression model is fitted on the reduced space representations. The hyper-parameters used are the same as the default ones from the sklearn implementation, %\footnote{\href{https://scikit-learn.org/stable/modules/generated/sklearn.linear_model.LogisticRegression.html}{\textcolor{blue}{https://scikit-learn.org/stable/modules/generated/sklearn.linear\_model.LogisticRegression.html}}}, 
except for the penalty parameter which is set to None, to ensure that the model performance depends solely on the quality of the dimensionality reduction rather than regularization effects.

\begin{figure}[!ht]
    \centering
    \includegraphics[width=1\linewidth]{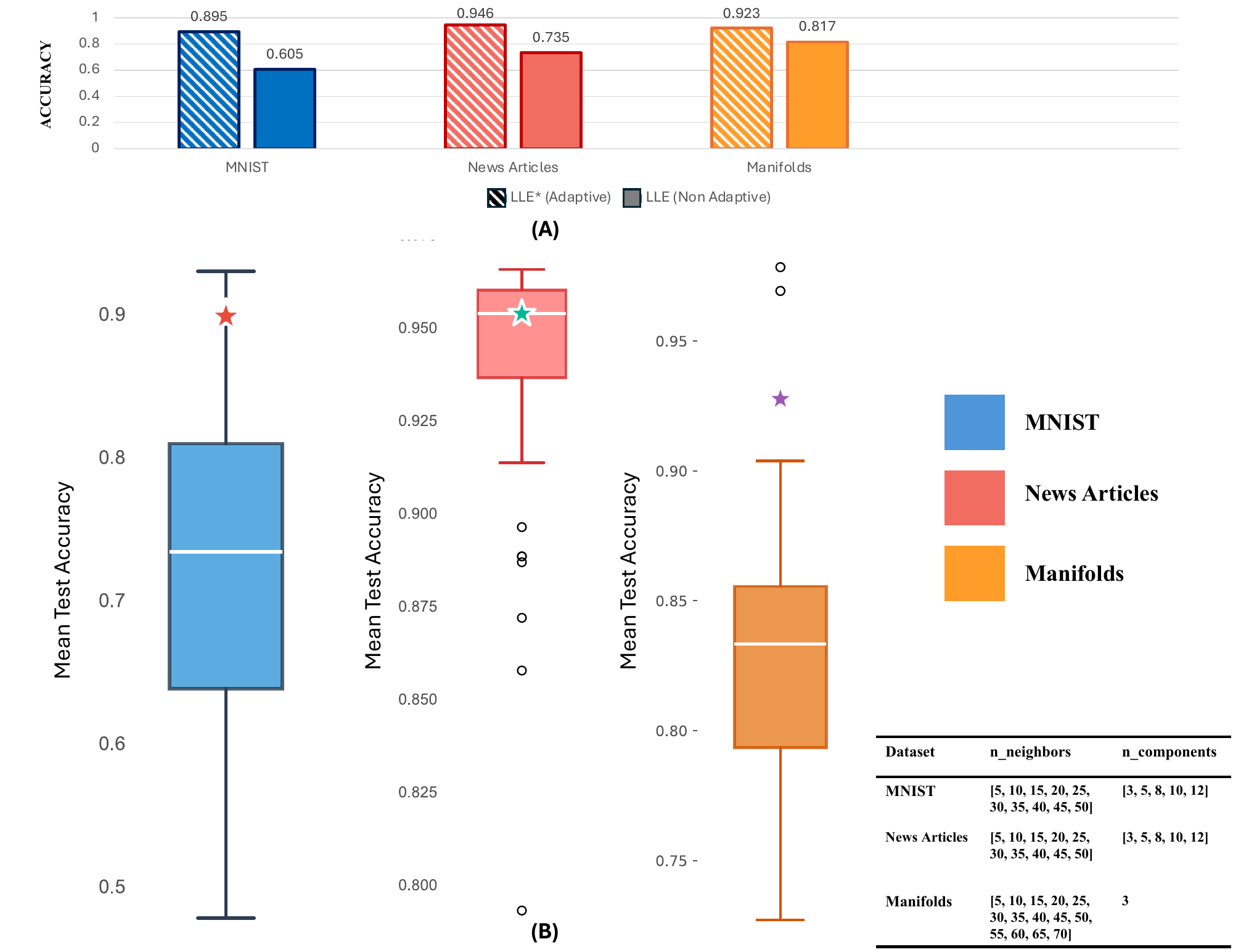}
\caption{\textbf{(A)} The results of the accuracy score for the datasets considered are shown for different configurations of LLE, adaptive (LLE$^*$) and non-adaptive (LLE). \textbf{(B)} Boxplots of accuracy scores comparing adaptive and non-adaptive dimensionality reduction approaches across MNIST, News Articles, and Manifolds datasets. The Boxplots show the distribution of scores from the hyper-parameter grid search, while the stars represent the adaptive method results. Results are averaged using an 80\%-20\% holdout approach over three different runs. The table on the bottom right shows the hyper-parameter grid used for the non-adaptive approach, exploring different combinations of n\_neighbours and n\_components values.}
\label{LLE}
\end{figure}

In Figure \ref{LLE} \textbf{(A)}, we report the results for LLE$^*$, and LLE implemented in sklearn%\footnote{\href{https://scikit-learn.org/dev/modules/generated/sklearn.manifold.LocallyLinearEmbedding.html}{\textcolor{blue}{https://scikit-learn.org/dev/modules/generated/sklearn.manifold.LocallyLinearEmbedding.html}}}
.
We observe that LLE$^*$ leads to the best results for all the datasets considered. Compared to the default sklearn implementation, which can be used as an easy-to-go choice, our adaptive approach achieves significantly better results in terms of accuracy, without any hyper-parameter tuning.

In order to extend the comparison between LLE$^*$ representations and their non-adaptive counterparts, we compare $\text{LLE}^*$ with LLE by varying the number of components and neighbours, exploring all possible combinations in a grid-search-like approach. In Figure \ref{LLE} \textbf{(B)}, we show the boxplots of accuracy scores based on different hyper-parameter-value combinations explored (shown in the bottom right table in Figure \ref{LLE} \textbf{(B)}). We observe that the use of $\text{LLE}^*$ approach allows us to achieve among the best results for the MNIST and Manifolds datasets, while for News Articles, the results are very close to the median, as they are already high. Also in this case, we empirically observe how the use of an adaptive approach consistently leads to robust and competitive results without the need for hyper-parameter tuning.

\section{Discussion}
We introduce a general framework for hyper-parameter tuning in non-linear NN-based dimensionality reduction methods, where the selection of the neighbourhood structure and the dimension of the projection space are often arbitrary. Our method is based on intrinsic dimension estimation and locally adaptive uniform neighbourhood selection, which are performed jointly. We show that our approach leads to a significant improvement in the performance and graphical visualisation of LLE, a very popular dimensionality reduction method in statistics and machine learning, whenever employed in unsupervised and supervised learning tasks on both simulated and real benchmark datasets. As a proof of concept, we also extend our approach to Spectral Clustering and UMAP, achieving excellent results in an unsupervised context. The framework can be naturally extended to all dimensionality reduction methods that require neighbourhood structure (n\_neighbours) and target space dimensionality (n\_components) as input to compute the projection. More broadly, the locally adaptive $k^*$-based paradigm can be extended to all unsupervised, semi-supervised, and supervised algorithms that leverage neighbourhood relationships (i.e., DBSCAN, Label Propagation, K-Nearest Neighbours, etc.).
\label{sec:disc}

\section{Acknowledgements}
All authors acknowledge financial support from the SNSF (Swiss National Science Foundation) grant  200557.

\bibliographystyle{apalike}
\bibliography{bib}

\end{document}